\documentclass[runningheads]{llncs}
\usepackage{graphicx}
\newcommand{\cmt}[1]{}
\usepackage[flushleft]{threeparttable}
\usepackage{hyperref}
\hypersetup{
  colorlinks   = true, 
  urlcolor     = black, 
  linkcolor    = red, 
  citecolor = blue 
}
\pdfoutput=1
\begin{document}

\title{Visual Semantic Re-ranker for Text Spotting}

\author{Ahmed Sabir\inst{1}\ 
Francesc Moreno-Noguer\inst{2} \and Llu\'{\i}s Padr\'o\inst{1}}

\institute{TALP Research Center, Universitat Polit\`ecnica de Catalunya, Barcelona, Spain \and
Institut de Rob\`otica i Inform\`atica Industrial (CSIC-UPC), Barcelona, Spain
\email{asabir@cs.upc.edu, fmoreno@iri.upc.edu, padro@cs.upc.edu}}

\maketitle              

\begin{abstract}

Many current state-of-the-art methods for text recognition are based on purely local information and ignore the semantic correlation between text and its surrounding visual context. In this paper, we propose a post-processing approach to improve the accuracy of text spotting by using the semantic relation between the text and the scene. We initially rely on an off-the-shelf deep neural network that provides a series of text hypotheses for each input image. These text hypotheses are then re-ranked using the semantic relatedness with the object in the image. As a result of this combination, the performance of the original network is boosted with a very low computational cost. The proposed framework can be used as a drop-in complement for any text-spotting algorithm that outputs a ranking of word hypotheses. We validate our approach on ICDAR'17 shared task dataset.   
\end{abstract}
\section{Introduction}
\label{sec:intro}

Machine reading has shown a remarkable progress in  Optical Character Recognition systems (OCR). However, the success of most OCR systems is restricted to simple-background and properly aligned documents. However, text in many real images  is affected by a number of artifacts including  partial occlusion, distorted perspective and complex backgrounds. For this reason, OCR systems able to read text in the wild are required, problem referred to  as \textit{Text Spotting}. However, while state-of-the-art computer vision algorithms have shown remarkable results in   recognizing object instances in these images, understanding and recognizing the included text in a robust manner is far from being considered a solved problem.  

Text spotting pipelines address the end-to-end problem of  detecting and recognizing text in unrestricted images (traffic signs, advertisements, brands in clothing, etc.). The problem is usually split in two stages: 1) \textit{text detection stage}, to estimate the bounding box around the candidate word in the image and 2) \textit{text recognition stage}, to identify the text inside the bounding boxes. In this paper we focus on the second stage, by means of a simple but efficient  post-processing approach built upon  Natural Language Processing (NLP) techniques. 

There exist two main approaches to perform text recognition in the wild. First,  lexicon-based methods, where the system learns to recognize words in a pre-defined dictionary \cite{jaderberg2016reading,wang2011end,almazan2014word}. Second, lexicon free, unconstrained recognition methods, that aim at predicting character sequences \cite{bissacco2013photoocr,shi2017end,ghosh2017visual}.

Most recent state-of-the arts systems are deep learning based approaches \cite{bissacco2013photoocr,ghosh2017visual,jaderberg2016reading,shi2017end}, which however, have some limitations: Lexicon-based approaches need a large dictionary to perform the final recognition. Thus, their accuracy will depend on the quality and coverage of this lexicon, which makes this approach unpractical for real world applications where the domain may be different to that the system was trained on. On the other hand, lexicon-free recognition methods rely on sequence models to predict character sequences, and thus they may generate likely sentences that do not correspond to actual language words. In both cases, these techniques rely on the availability of large datasets to train and validate, which may not be always available for the target domain.

In this work we propose a post-processing approach via a  \textit{visual context re-ranker} to overcome these limitations. Our approach uses visual prior to re-rank the candidate words based on the semantic relation between the scene text and its environmental \textit{visual} context. Thus, the visual re-ranker can be applied to any of both methods, the huge dictionary in case there is a lexicon, or unconstrained recognition such as character prediction sequence, to re-rank most probable word biased by its visual information. 

The work of \cite{patel2016dynamic} also uses  visual prior information to improve text spotting results, through a new lexicon built with Latent Dirichlet Allocation (LDA) \cite{blei2003latent}. The topic modeling learns the relation between text and images. However, this approach relies on captions describing the images rather than using the main keywords semantically related to the images to generate the lexicon re-ranking. Thus, the lexicon generation can be inaccurate in some cases due to the short length of captions. In this work we consider a direct semantic relation between scene text and its visual information. Also, unlike \cite{patel2016dynamic} that only uses visual information over word frequency count to re-rank the  most probable word, our approach combines both methods by leveraging also on a frequency count based language model.

Our main contributions therefore include several post-processing methods based on NLP techniques such as word frequencies and semantic relatedness which are typically used in pure NLP problems but less common in computer vision. We show that by introducing a candidate re-ranker based on word frequencies and semantic distances between candidate words and objects in the image, we can improve the performance of an off-the-shelf deep neural network without the need to perform additional training or tuning.
In addition, thanks to the inclusion of the unigram probabilities, we overcome the baseline limitation of false detection of short words of~\cite{jaderberg2016reading,ghosh2017visual}.

\begin{figure}[t!]
\centering 

\includegraphics[width=0.8\textwidth]{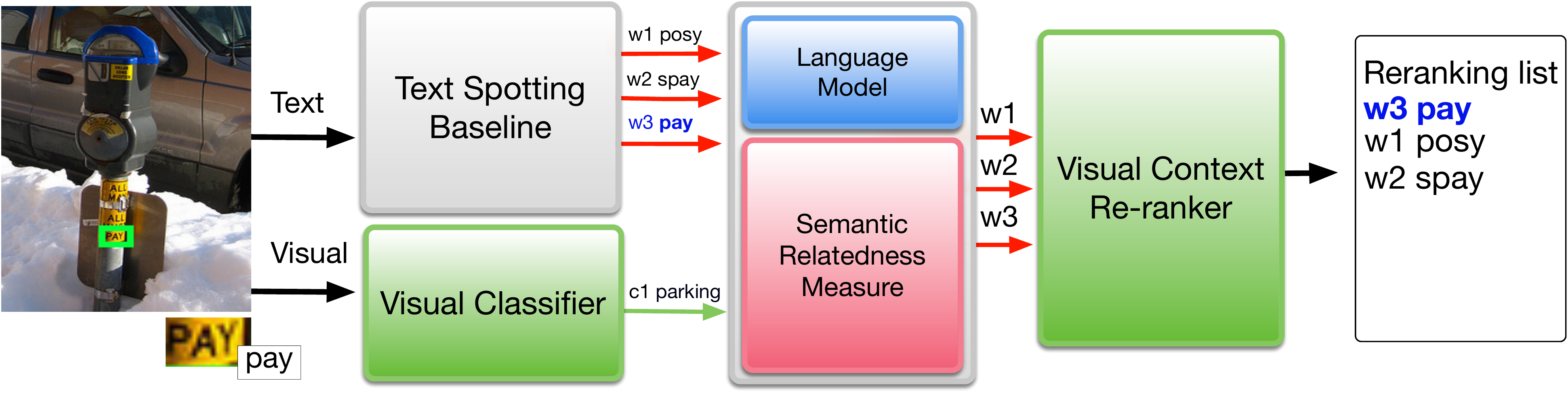}
\vspace{-2mm}

\caption{Scheme of the proposed visual context information pipeline integration into the text spotting system. Our approach uses the language model and a semantic relatedness measure to re-rank the word hypotheses. In this example, the re-ranked word \textit{pay} is semantically related with the visual \textit{parking}.}
 \label{lm-model1} 
 \end{figure}

The rest of the paper is organized as follows: Section  \ref{Baseline} describes our proposed pipeline. Section \ref{Visual context bias information} introduces the external prior knowledge we use, namely a visual context information. Section \ref{Experiments result} presents experimental validation of our approach on a variety of publicly available standard datasets. Finally, Section \ref{Conclusion} summarizes the work and discusses specifies future  work.

\section{General Approach and Baseline System}
\label{Baseline}

There exist two main classes of recognition approaches: 1) character based methods that rely on a single character classifier plus some kind of sequence modeling (e.g. n-gram models or  LSTMs), and 2) lexicon-based approaches that aim to classify the image as a whole word.

In both cases, the system can be configured to predict the $k$ most likely words given the input image. Our approach focuses on re-ranking that list using semantic relatedness with objects in the image (or \texttt{visual context}) in which the text was located.

We used two different  off-the-shelf baseline models: First, a CNN \cite{jaderberg2016reading} with fixed lexicon based recognition. It uses a fixed dictionary of around 90K words. Second, we considered a LSTM architecture with a visual attention model \cite{ghosh2017visual}. The LSTM generates the final output words as character sequences, without relying on any lexicon. Both models are trained on a synthetic dataset \cite{jaderberg2014synthetic}. The output of both models is a vector of softmax probabilities for candidate words. We next describe in more  detail these components.

Let us denote the initial probabilities of the $k$ most likely words $w$ produced by each of the \textit{baselines} (BL) \cite{jaderberg2016reading,ghosh2017visual} by: 
\begin{equation} 
{P_0(w) = p(w|\textrm{BL})}
\end{equation}
We introduce a Unigram Language Model (ULM) as preliminary stage for the visual re-ranker.  The probabilities of the unigram model are estimated on the {\textit{Opensubtitles}}\footnote{\href{https://www.opensubtitles.org}{\textcolor{black}{https://www.opensubtitles.org}}} \cite{tiedemann2009news} and {\textit{Google book n-gram}}\footnote{\href{https://books.google.com/ngrams}{\textcolor{black}{https://books.google.com/ngrams}}} text corpora. The main goal of ULM is to increase the probability of the most common words proposed by the baseline. The output of this re-ranker is then used as input for a second re-ranker based on visual context information. The ULM should be useful to 1) enhance the baseline recognition output 2) and to ensure each output is a valid word before performing the visual semantic re-ranking stage.

\section{Visual Context Information }
\label{Visual context bias information}

The main source of semantic information we use to re-rank the BL output is the visual context information, i.e. the semantic relation between the candidate words and the objects in the image.
We use a pre-trained object classifier to detect the objects in the image and we devise a strategy to reward candidate words that are semantically related to them. As shown in Figure \ref{lm-model1} the top position of the re-ranking yields \textit{pay} as the most semantically related with the object detected in the image \textit{parking}. 
Next, we describe two different schemes to compute this relatedness.

\subsection{Visual Classifier}
In order to exploit and extract the image context information we use state-of-the-art object classifiers. We considered two pre-trained CNN classifiers: ResNet \cite{He2015} and GoogLeNet \cite{szegedy2015going}. \cmt{However, we used only resnet with 152 layers network, due to the superior accuracy over GoogLnet.}The output of these classifiers is a $1000$-dimensional vector with the probabilities of $1000$ object instances. In this work we only consider the most likely object of the context. We set a threshold to filter out the probability predictions when the object classifier is not confident enough.

\subsection{Semantic Relatedness with Word Embedding (SWE)}

Once the objects in the image have been detected, we compute their semantic relatedness with the candidate words based on their word-embeddings \cite{mikolov2013distributed}. Specifically, let us denote by $\vec{w}$ and $\vec{c}$ the word-embeddings of a candidate word $w$ and the most likely object $c$ detected in the image. We then compute their similarity using the cosine of the embeddings:
\begin{equation} 
sim(w, c) = \frac{\vec{w}\cdot \vec{c}}{|\vec{w}|\cdot|\vec{c}|}\;
\end{equation}

We next  convert the similarity score to a probability value, in order to integrate it into our re-ranking model. Following \cite{blok2003probability}, we compute the conditional probability from similarity as:
\begin{equation}
\label{eq:sim}
P_{SWE}(w\vert c)=P(w)^\alpha \mathrm{~~~~~where~} \alpha=\left({\textstyle\frac{1-sim(w,c)}{1+sim(w,c)}}\right)^{1-P(c)\;}
\end{equation}
$P(w)$ is the probability of the word in general language (obtained from the unigram model), and $P(c)$ is the probability of the context object (obtained from the object classifier).

Once we have the probability of a candidate word conditioned to the visual context objects, we can define $P(w|\textrm{SWE})=P_{SWE}(w|c)$ and use it to re-rank the output of the BL :
\begin{equation}
P_1(w)=P(w\vert \textrm{BL})\times P(w\vert \textrm{SWE})
\end{equation}
Note that the frequency information from the ULM is already included in $P(w|\textrm{SWE})$, so this step will rely on the output of the language model.

\subsection{Estimating Relatedness from Training Data Probabilities (TDP)}
A second possibility to compute semantic relatedness is to estimate it from training data. This should overcome the word embedding limitation when the candidate word and the image objects are not semantically related in general text, but occurred in the real world. For instance, a tennis sponsor (watch brand) \textit{rolex} and the object \textit{racket}, have no semantic relation according to the word embedding model, but they are found paired multiple times in the training dataset, which implies they do have a relation. 
To encode this relation, we use training data to estimate the conditional probability of a word $w$ given that object $c$ appears in the image:
\begin{equation}
\small P_{TDP}(w\vert c)\;=\;\frac{count(w,c)}{count(c)} \;
\end{equation}
where $count(w,c)$ is the number of training images where $w$ appears as the gold standard annotation for recognized text, and the object classifier detects object$c$ in the image. Similarly, $count(c)$ is the number of training images where the object classifier detects object $c$. 

 We combined this re-ranker with SWE as:
\begin{equation}
P_2(w)=P(w\vert \textrm{BL})\times P(w\vert \textrm{SWE}) \times  P(w\vert \textrm{TDP})
\end{equation}

\subsection{Semantic Relatedness with Word Embedding (revisited) (TWE)}
This re-ranker builds upon a word embedding, as the SWE re-ranker above, but the embeddings are learnt from the training dataset (considering two-word ``sentences'': the target word and the object in the image). The embeddings can be computed from scratch, using only the training dataset information (TWE). 

In this case, we convert the similarity produced by the embeddings to probabilities using:
\begin{equation}
\label{eq:tanh}
\small  \small P_{TWE}(w\vert c)=\frac{\tanh(sim(w,c))+1}{2 P(c)}
\end{equation}
Note that this re-ranker does not take into account word frequency information as in the case of the SWE re-ranker. Also, we add this re-ranker as : 
\begin{equation}
P_3(w)=P(w\vert \textrm{BL})\times P(w\vert \textrm{TWE}) \times  P(w\vert \textrm{TDP})
\end{equation}

\section{Experiments and Results}
\label{Experiments result}

In this section we evaluate the performance of the proposed approaches on the 
{\bf ICDAR-2017-Task3 (end-to-end)} dataset  \cite{veit2016coco}. This dataset is based on Microsoft COCO \cite{lin2014microsoft} (Common Objects in Context), which consists of 63,686 images, and 173,589 text instances (annotations of the images). COCO-text was not collected with text recognition in mind, therefore, not all images contain textual annotations. The \textit{ICDAR-2017 Task3} aims for end-to-end text spotting (i.e. both detection and recognition). Thus, this dataset includes full images, and the texts in them may appear rotated, distorted, or partially occluded. Since we focus only on text recognition, we use the ground truth detection as a golden detector to extract the bounding boxes from the full image. The dataset consists of 43,686 full images with 145,859 text instances for training, and 10,000 images with 27,550 text instances for validation.

\cmt{This dataset is much larger than previous ICDAR datasets, it enables  training end-to-end deep architectures for text detection and recognition} 

\cmt{\subsection{Implementation details}}
\subsection{Preliminaries} 
\label{preliminaries}

For evaluation, we used a more restrictive protocol than the standard proposed by \cite{wang2011end} and adopted in most state-of-the-art benchmarks, which does not consider words with less than three characters or with non-alphanumerical characters. This protocol was introduced to overcome the false positives on short words that most current state-of-the-art struggle with, including our Baseline. However, we overcome this limitation by adopting a language model before the visual re-ranker. Thus, we consider all cases in the dataset, and words with less than three characters are also evaluated. 

In all cases, we use two pre-trained deep models, CNN \cite{jaderberg2016reading} and LSTM  \cite{ghosh2017visual} as a baseline (BL) to extract the initial list of word hypotheses. Since these BLs need to be fed with the cropped words, when evaluating on the ICDAR-2017-Task3 dataset we will use the ground truth bounding boxes of the words.

\label{experimnet1}

\subsection{Experiment with Visual Context Information}
\label{experiment2}
The main contribution of this paper consists in  re-ranking the $k$ most likely hypotheses using the visual context information. Thus, we use ICDAR-2017-Task3 dataset to evaluate our approach, re-ranking the baseline output using the semantic relation between the spotted text in the image and its visual context.  
 
We extract the $k=5..9$ most likely words --and their probabilities-- from the baselines. The first baseline is a CNN \cite{jaderberg2016reading} with fixed-lexicon recognition, which is not able to recognize any word outside its dictionary. We present three different accuracy metrics: 1) \textit{full} columns correspond to the accuracy on the whole dataset, 2) \textit{dict} columns correspond to the accuracy over the cases where the target word is among the 90K-words of the CNN dictionary (which correspond to 43.3\% of the whole dataset) and finally 3) \textit{list} shows the accuracy over the cases where the right word was in the $k$-best list output by the baseline. The second  baseline we consider is an LSTM \cite{ghosh2017visual} with visual soft-attention mechanism, using unconstrained text recognition approach without relying on a lexicon.

Both baselines work on cropped words, we do not evaluate the whole end-to-end system but only the influence of adding external knowledge. We used ground-truth bounding boxes as input to the BL. Thus, the whole image is used as input to the object classifier.

In order to extract the visual context information  we considered two different pre-trained state-of-the-art object classifiers: ResNet  \cite{He2015} and GoogLeNet \cite{szegedy2015going}, both able to detect pre-defined list of 1,000 object classes.
In this experiment we re-rank the baseline $k$-best hypotheses based on their relatedness with the objects in the image. We try two approaches for that: 1) semantic similarity computed using word embeddings \cite{mikolov2013distributed} and 2) correlation based on co-ocurrence of text and image object in the training data. 

\begin{table}[t!]
\small 
\centering
\caption{Results of re-ranking the $k$-best ($k=5,9$) hypotheses of the baselines on ICDAR-2017-Task3 dataset (\%)} 
\label{table_1} 
 \begin{tabular}{|c| c c c|c c c|c c | c c |} 
 \hline
 \textbf{Model}  & \multicolumn{6}{c|}{\textbf{CNN}}   & \multicolumn{4}{c|}{\textbf{LSTM}}  \\ 
  & \multicolumn{3}{c|}{$k=5$} & \multicolumn{3}{c|}{$k=9$} &  \multicolumn{2}{c|}{$k=5$} &   \multicolumn{2}{c|}{$k=9$}   \\ 
  & \textit{full} & \textit{dict} & \textit{list} &  \textit{full} &  \textit{dict} &\textit{list} &   \textit{full}  & \textit{list} &  \textit{full} &  \textit{list}\\ 
 \hline\hline
   Baseline  & \multicolumn{6}{c|}{\textbf{full: 21.1 dict: 58.6}}   & \multicolumn{4}{c|}{\textbf{full: 18.7}}  \\
   \hline 
 \hline

SWE         &   22.5    &  62.5  & 80.6  & 22.6   &   62.8      & 70.4        & 19.5  &  70.1   & 19.9          &  62.3 \\
SWE+TDP      &   22.7    &  63.1  & 81.6  & 21.4  &   59.5      & 66.6        & 19.5  &  70.1   & 20.0          &  62.6  \\
TDP+TWE     &   22.9    &  63.8  & 82.2  &  \textbf{23.0}   & \textbf{64.0}   & \textbf{71.6}  & 20.0   &  73.0 & \textbf{20.8} &  \textbf{65.2}  \\ 
\small SWE+TDP+TWE   &   22.6    &  62.9  & 81.0  & 22.5 &   62.5 & 70.0  & 20.1  &  72.3 & 20.3          &  63.6   \\

 \hline
 \end{tabular}
\end{table}

First, we re-rank the words based on their embedding-based semantic relatedness with the image visual context (SWE). For instance,  the semantic similarity between a visual \textit{street} and text \textit{way} in a signboard.

Secondly, we use the training dataset to compute the conditional probabilities between text image and object in the image happen together (TDP). As shown in Table~\ref{table_1} the LSTM accuracy improved up to 1.3\%, and the fixed-lexicon CNN accuracy is boosted up to 1.6\% on \textit {full} dataset and 2.5\% \textit{dictionary}. 

Finally, we trained a word embedding model using the training dataset (TWE). Due to the dataset is too small, we use skip-gram model with one window, and without any word filtering. In addition, we initialized the model weight with the baseline (SWE) that trained on general text. The result is 300-dimension vector for about 10K words. The result in  Table~\ref{table_1} shows that (TWE) outperform the accuracy of SWE model that trained on general text. The result in Table ~\ref{table_1} shows that the combination model TDP+TWE also significantly boost the accuracy up to 5.4\% \textit{dictionary} and 1.9\% \textit{all}. Also, the second baseline LSTM accuracy boosted up to 2.1\%. Not to mention that TDP+TWE model only rely on the visual context information, computed by Equation \ref{eq:tanh}.    

\section{Conclusion}
\label{Conclusion}

In this paper, we have proposed a simple visual context re-ranker as post-processing approach to a text spotting system. We have shown that the accuracy of two different architecture state-of-the-art, lexicon based and lexicon free, deep networks can be improved to 2 points on standard benchmark. In the future work, we plan to explore more visual context such as multiple objects and information from the scene.


\bibliography{references}

 \bibliographystyle{splncs04}

\end{document}